\documentclass{article}

\usepackage[preprint]{neurips_2026}
\makeatletter
\renewcommand{\@notice}{} 
\makeatother

\usepackage[utf8]{inputenc} 
\usepackage[T1]{fontenc}    
\usepackage{hyperref}       
\hypersetup{pdftitle={Timestral}}
\usepackage{url}            
\usepackage{booktabs}       
\usepackage{amsfonts}       
\usepackage{nicefrac}       
\usepackage{microtype}      
\usepackage{xcolor}         
\usepackage{graphicx}
\usepackage{placeins}
\usepackage{subcaption}
\usepackage{comment}
\usepackage[nohyperlinks]{acronym}
\usepackage{xspace}

\usepackage[capitalise,noabbrev,nameinlink]{cleveref}

\newacro{tsfm}[TSFM]{time series foundation model}
\newacroplural{tsfm}[TSFMs]{time series foundation models}

\newacro{cpm}[CPM]{Contiguous Patch Masking}
\newacro{dhf}[DHF]{Decoupled Horizon Forecasting}
\newacro{ar}[AR]{autoregressive}
\newacro{scm}[SCM]{structural causal model}
\newacro{dag}[DAG]{directed acyclic graph}
\newacro{rnn}[RNN]{recurrent neural network}
\newacro{mlp}[MLP]{multilayer perceptron}

\newcommand{\timestral}{Timestral\xspace}
\newcommand{\timestralSize}[1]{Timestral-#1\xspace}
\newcommand{\totoTwo}{Toto-2.0\xspace}
\newcommand{\totoTwoSize}[1]{Toto-2.0-#1\xspace}
\newcommand{\chronosTwo}{Chronos-2\xspace}
\newcommand{\timesfmThree}{TimesFM-3\xspace}
\newcommand{\falconTwo}{Falcon-2.0\xspace}
\newcommand{\tirexTwo}{TiRex-2\xspace}
\newcommand{\tzeroGamma}{t0-gamma\xspace}
\newacro{mase}[MASE]{mean absolute scaled error}
\newacro{crps}[CRPS]{continuous ranked probability score}
\newacro{sql}[SQL]{scaled quantile loss}
\newacro{nmase}[nMASE]{normalised \acs{mase}}
\newacro{ncrps}[nCRPS]{normalised \acs{crps}}
\newacro{ppd}[PPD]{posterior predictive distribution}
\newacro{pfn}[PFN]{prior-data fitted network}
\newacroplural{pfn}[PFNs]{prior-data fitted networks}

\newcommand{\targetvalue}{y}
\newcommand{\covariatevalue}{x}

\newcommand{\patchemb}{\mathbf{v}}
\newcommand{\predtoken}{\mathbf{p}}
\newcommand{\maskedpatchemb}{\mathbf{m}}
\newcommand{\finalpatchemb}{\mathbf{h}}

\newcommand{\forecastvalue}{\hat{\mathbf{\targetvalue}}}
\newcommand{\covariateemb}{\mathbf{c}}
\newcommand{\pastcovemb}{\covariateemb^{\mathrm{past}}}
\newcommand{\knowncovemb}{\covariateemb^{\mathrm{fut}}}

\newcommand{\modelfigurescale}{1}

\title{Timestral}

\author{} 

\begin{document}

\maketitle

\vspace{-0.1in}
\begin{center}
\vspace{-45pt}
\centering
\includegraphics[width=0.8\linewidth,keepaspectratio]{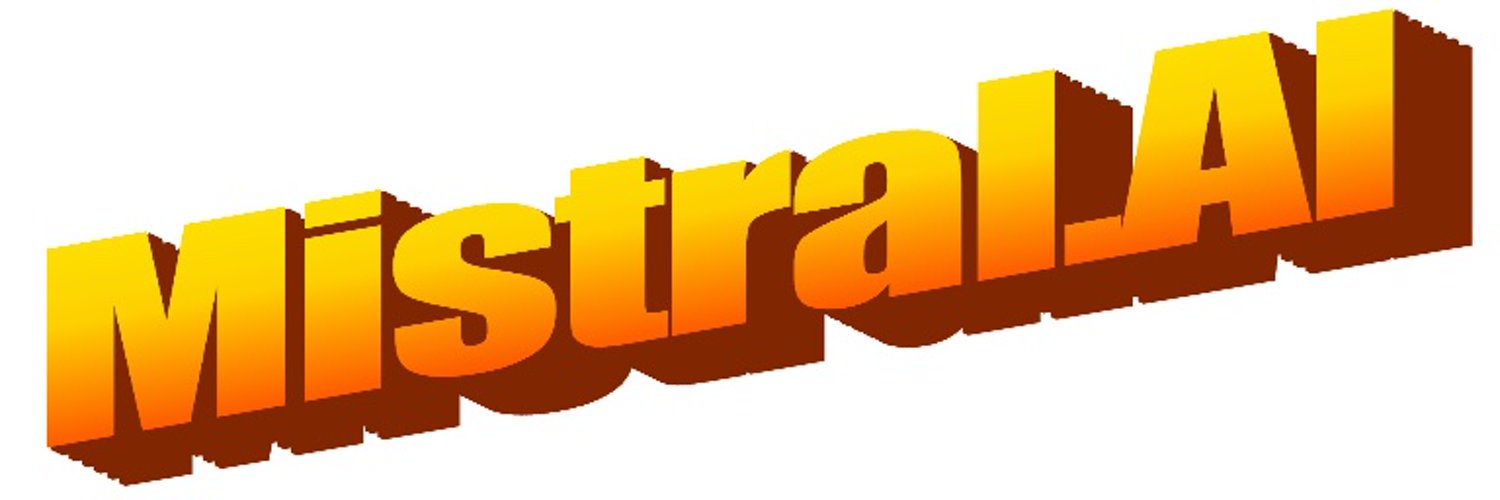}
\end{center}

\begin{abstract}
  We introduce \textbf{Timestral}, a family of four time series foundation models for probabilistic forecasting, with 5M, 28M, 123M, and 209M parameters. 
  \timestral sets a new state of the art on GIFT-Eval, TIME, and fev-bench, where \timestralSize{209M} and \timestralSize{123M} take the top two places on every metric.
  We achieve this with three ingredients: Decoupled Horizon Forecasting (DHF), which separates prediction tokens from the causally encoded context, so that longer horizons are trained without masking context and each observed value is both a forecast target and context for later forecasts; a covariate framework that natively supports known-future covariates within a causal architecture; and a two-stage pipeline that pretrains exclusively on a diverse, large-scale synthetic corpus before post-training on a mix of real and synthetic data with extended context. 
  This report describes the model design and its benchmark results.
  Weights for all four models are available on Hugging Face\footnote{\url{https://huggingface.co/collections/mistralai/timestral}} and inference code on GitHub\footnote{\url{https://github.com/mistralai/timestral}}.
\end{abstract}

\begin{center}
  \hypersetup{pdfborder={0 0 0}}
  \textbf{Webpage:} \href{https://mistral.ai/news/timestral}{\textcolor{blue}{\nolinkurl{https://mistral.ai/news/timestral}}}
\end{center}
\acresetall

\section{Introduction}

\suppressfloats[t]
\begin{figure}[t]
    \centering
    \includegraphics[width=1.0\linewidth]{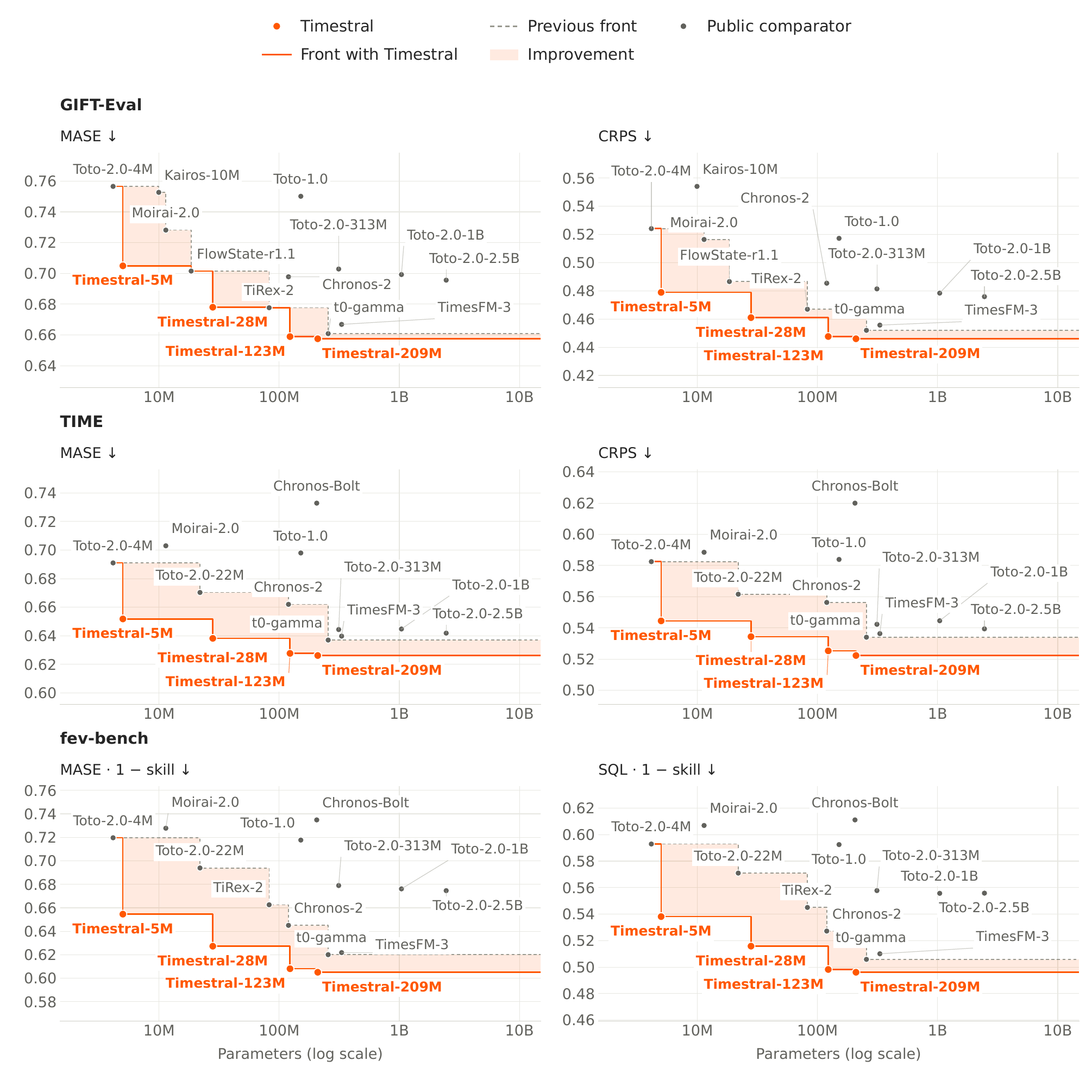}
    \caption{Forecast error against parameter count on GIFT-Eval, TIME, and fev-bench (lower is better; fev-bench skill scores are shown as one minus the skill score).
    The dashed and solid curves show the Pareto fronts without and with \timestral, respectively.
    All four \timestral models lie on the Pareto front across the six panels.}
    \label{fig:pareto}
\end{figure}

\Acp{tsfm} enable forecasting across datasets and domains without retraining. By training on large collections of real and synthetic time series, these models provide a shared alternative to fitting a separate model for each forecasting problem~\citep{ansari2024chronos, das2024timesfm}. 
Recent advances have improved forecast accuracy, demonstrated gains from scaling, and extended support to multivariate series and external covariates~\citep{khwaja2026toto2, ansari2025chronos2, podest2026tirex2, meyer2026t0timeseriesfoundationmodel}. 
These developments place increasing importance on how models use the information available at each forecast origin and on the training distributions that support this use.

Bidirectional encoder-only models such as \chronosTwo and \falconTwo~\citep{ansari2025chronos2, xia2026orbit} sample a context length and a horizon length for each training example, so each forward pass covers a single combination of the two.
Causal decoder-only models instead forecast from every position of a sequence and thus cover many context lengths in one forward pass.
To also cover multiple horizon lengths, recent causal models are trained to forecast several patches ahead directly rather than autoregressively.
\Ac{cpm}, used in TiRex, \totoTwo, and \timesfmThree~\citep{auer2025tirex, khwaja2026toto2, jain2026timesfm3}, achieves this by masking spans of the input sequence.
The masked observations, however, are then also unavailable as context for later forecasts in the same training sequence, coupling the horizon distribution to full-patch context masking.
Known-future covariates pose a second challenge: their future values must be accessible to forecasts, while future target values must remain hidden.

We introduce \textbf{Timestral}, a family of four foundation models for probabilistic univariate, multivariate, and covariate-informed time series forecasting, with 5M, 28M, 123M, and 209M parameters. 
All four models share the same architecture and training recipe, offering a range of trade-offs between forecast accuracy and inference cost. 
\timestral addresses the coupling of horizons and context masking through \ac{dhf}, which separates causal context representations from learned prediction tokens, and the covariate challenge through a role-aware covariate framework that integrates known-future information while preserving target causality. 
The models are pretrained on a synthetic corpus of univariate signals and structured multivariate dependencies, then post-trained on a mixture of real and synthetic data with extended context. 
\timestralSize{209M} sets a new state of the art in forecasting performance on GIFT-Eval, TIME, and fev-bench.
As shown in \cref{fig:pareto}, all four \timestral models lie on the Pareto front for all benchmark metrics.
\timestralSize{123M} already outperforms every other model shown, including models with billions of parameters.

This report presents the released \timestral models, their benchmark results, and the main design choices behind them:
\begin{itemize}
    \item \textbf{\acl*{dhf}.} We introduce \ac{dhf}, a forecasting formulation that separates missing observations from prediction tokens. 
    \Ac{dhf} preserves observed values as context for later forecast origins within the same training sequence, allowing multiple forecast blocks to be supervised while controlling horizon sampling independently of artificial context masking.
    \item \textbf{Covariate conditioning within \acs*{dhf}.} We integrate past-only and known-future covariates through explicit series roles and asymmetric attention. 
    This enables prediction tokens to use known-future information at multiple forecast origins while preserving causality over target values.
    \item \textbf{A model family spanning accuracy and inference cost.} We release four model sizes, from 5M to 209M parameters. 
    The 209M model ranks first on all reported metrics among the models compared on GIFT-Eval~\citep{aksu2024giftplus}, TIME~\citep{qiao2026time}, and fev-bench~\citep{shchur2025fevbench}. 
\end{itemize}

\section{Background}
\label{sec:background}

We first describe \acp{tsfm} within the framework of \acp{pfn}, which interprets forecasting as amortised posterior inference (\cref{sec:background:pfn}).
We then review how current \acp{tsfm} model dependencies across variates and over time, and the drawbacks that motivate the design of \timestral (\cref{sec:background:modeling}).

\subsection{Time Series Foundation Models as Prior-Data Fitted Networks}
\label{sec:background:pfn}

Synthetic data has become a standard ingredient of \ac{tsfm} training, and its share has generally grown.
TimesFM mixes 20\% synthetic series built from ARMA processes, seasonal patterns, trends, and steps \citep{das2024timesfm}; Chronos samples 10\% of its training series from KernelSynth, a Gaussian-process generator \citep{ansari2024chronos}; and TiRex draws about a third of its training series from an extended KernelSynth generator \citep{auer2025tirex}.
More recent models rely on synthetic data more heavily: 57.5\% of \totoTwo's pretraining mix is generated with TempoPFN's generators \citep{khwaja2026toto2}, and \chronosTwo learns its multivariate and covariate capabilities from synthetic data only, with a variant trained without any real data performing only slightly worse than the full model \citep{ansari2025chronos2}.

At the other end, time series \acp{pfn} such as ForecastPFN, Mamba4Cast, TimePFN, and TempoPFN are trained exclusively on synthetic data \citep{dooley2023forecastpfn, bhethanabhotla2024mamba4cast, taga2025timepfn, moroshan2025tempopfn}.

\timestral follows this direction: it is pretrained exclusively on synthetic data, and a significant part of our effort went into designing the generating processes.
The \ac{pfn} framework provides a principled view of this choice, in which the generators define a prior over processes and forecasting becomes approximate Bayesian inference under that prior.
  
Given the observed history $\mathbf{\targetvalue}_{1:t}$ of the target variates (the context), the task is to produce forecasts $\forecastvalue_{t+1:t+H}$ for the next $H$ time steps.
From a Bayesian perspective, a probabilistic forecast should follow the \ac{ppd}.
If the series is generated by an unknown process $f$ drawn from a prior $p(f)$, the \ac{ppd} integrates over all processes consistent with the context,
\begin{equation}
p(\mathbf{\targetvalue}_{t+1:t+H} \mid \mathbf{\targetvalue}_{1:t})
= \int p(\mathbf{\targetvalue}_{t+1:t+H} \mid \mathbf{\targetvalue}_{1:t}, f)\,
p(f \mid \mathbf{\targetvalue}_{1:t})\, \mathrm{d}f.
\label{eq:background:ppd}
\end{equation}
It accounts for both the noise of a given process and the uncertainty about which process generated the context.
Covariates enter as further conditioning variables and are omitted for brevity.

\Acp{pfn} \citep{muller2022pfn} learn to approximate the \ac{ppd} during training: each training example draws a process $f \sim p(f)$ and a dataset from $f$, and the network predicts held-out values from the observed ones.
In tabular \acp{pfn} \citep{hollmann2023tabpfn, hollmann2025tabpfn}, $f$ can be a random \acl*{scm} that generates context rows of features and targets; the network then predicts the targets of further rows from their features.
In time series \acp{pfn}, $f$ generates a context $\mathbf{\targetvalue}_{1:t}$ (optionally with covariates) and the network predicts its continuation $\mathbf{\targetvalue}_{t+1:t+H}$ as in \cref{eq:background:ppd}.
When trained with the negative log-likelihood, the network's optimal prediction is the \ac{ppd}.
The network thus amortises posterior inference over $f$: instead of fitting a model to each new series, it infers from the context which processes are plausible and approximates the \ac{ppd} in a single forward pass.

\Acp{tsfm} differ from this setting in two respects but admit the same interpretation.
First, many recent \acp{tsfm} are trained with the pinball loss \citep{auer2025tirex, ansari2025chronos2, khwaja2026toto2}.
With the pinball loss, the optimal prediction is not the full \ac{ppd} but its quantiles \citep{koenker1978regression}: for each future time step and quantile level $\tau$, the network learns the $\tau$-quantile of the \ac{ppd} at that time step.
Second, their prior is not defined by synthetic generators alone: real series, viewed as draws from unknown processes, are also used, for example in continued training as in recent tabular \acp{pfn} \citep{garg2025realtabpfn}, so the full training data and its augmentations form the prior.

In this view, a \ac{tsfm} approximates the \ac{ppd} under the prior defined by its training data.
This prior determines which processes are plausible for a given context and thus how uncertain the forecasts are.
A prior with too much unpredictable variability widens forecasts even for regular series, whereas one with too little makes the model overconfident.
The composition of the training data is therefore a central design choice for \acp{tsfm}.

\subsection{Time Series Foundation Model Architectures}
\label{sec:background:modeling}

\paragraph{Patch-based forecasting.} Following PatchTST \citep{nie2023patchtst}, most modern \acp{tsfm} represent a time series as a sequence of non-overlapping patches \citep{das2024timesfm, woo2024moirai, ansari2025chronos2, khwaja2026toto2, jain2026timesfm3}.
Each patch $i$ of $P$ consecutive time steps of a target variate is mapped to a token embedding $\patchemb_i$, reducing the token sequence length by a factor of $P$.
A sequence-modelling backbone, such as a Transformer, state-space model, or \acl*{rnn}, processes these context tokens and computes patch-based representations $\finalpatchemb_j$ for the forecast horizon.
A prediction head maps these representations back to time-step resolution, producing forecasts for future patch $j$.
This is commonly a quantile head, which predicts conditional quantiles for each time step to represent forecast uncertainty \citep{ansari2025chronos2, khwaja2026toto2}.

\paragraph{Multivariate forecasting and covariates.} Multivariate forecasting allows predictions for each target variate to depend on both its own history and the histories of other target variates.
Covariates $\mathbf{\covariatevalue}_t$ provide additional inputs without themselves being forecast.
Past-only covariates (patch embeddings $\pastcovemb_i$) supply historical values, while known-future covariates (patch embeddings $\knowncovemb_i$) supply both historical values and values known in advance over the forecast horizon.
\totoTwo and \chronosTwo use cross-variate attention to exchange information between variates at the same patch position.
\totoTwo replaces temporal attention in its final Transformer block with variate attention, whereas \chronosTwo applies both temporal and variate attention in every Transformer block \citep{khwaja2026toto2, ansari2025chronos2}.
\chronosTwo supports both covariate types within its group-attention formulation.

\begin{figure}[tb]
    \centering
    \begin{subfigure}[t]{0.49\linewidth}
        \centering
        \includegraphics[scale=\modelfigurescale]{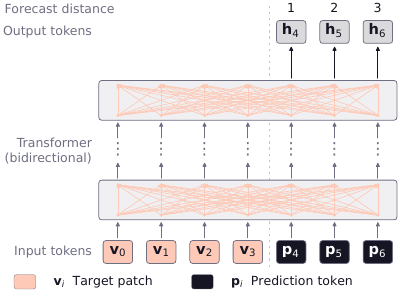}
        \caption{Bidirectional encoder-only model.}
        \label{fig:background:bidirectional_train}
    \end{subfigure}
    \hfill
    \begin{subfigure}[t]{0.49\linewidth}
        \centering
        \includegraphics[scale=\modelfigurescale]{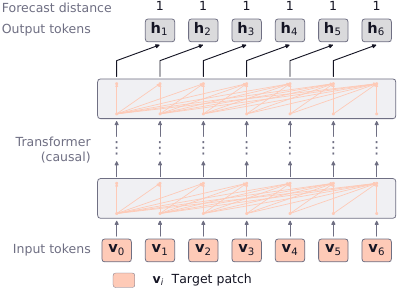}
        \caption{Causal decoder-only model.}
        \label{fig:background:causal_ar_train}
    \end{subfigure}
    \caption{Temporal information flow during training.
    Input tokens enter the Transformer after embedding; output tokens are final Transformer representations supplied to the prediction head.
    Only outputs used for forecasting are shown.
    The index of $\finalpatchemb_j$ denotes the forecast patch $j$, and numbers above it give forecast distances in patches.
    (a) Context tokens $\patchemb_0,\ldots,\patchemb_3$ and prediction tokens $\predtoken_4,\ldots,\predtoken_6$ interact bidirectionally, producing $\finalpatchemb_4,\finalpatchemb_5,\finalpatchemb_6$ at the future positions.
    (b) Causal attention computes $\finalpatchemb_{i+1}$ at position $i$ for one-patch-ahead forecasting.
    Normalisation, embedding, the prediction head, and denormalisation are omitted here and shown in \cref{fig:timestral:overview}.}
    \label{fig:background:bidir_enc_vs_causal_dec_train}
\end{figure}

\paragraph{Bidirectional temporal modelling.} Bidirectional encoder-only models such as \chronosTwo \citep{ansari2025chronos2} and \falconTwo \citep{xia2026orbit} process the observed context together with tokens at future prediction positions, without exposing future values of the target variates.
\cref{fig:background:bidirectional_train} illustrates this construction with context embeddings $\patchemb_0,\ldots,\patchemb_3$ and prediction tokens $\predtoken_4,\ldots,\predtoken_6$.
Bidirectional attention lets all positions exchange information; the prediction head maps the final representations at future positions to forecasts.
Training on forecasts starting at different points in a series generally requires encoding each context separately, because bidirectional attention lets later observations influence earlier context representations.

\paragraph{Causal temporal modelling.} Causal temporal modelling restricts each context representation to values of target variates and past-only covariates at the current and preceding patch positions.
Known-future covariate values may additionally be used beyond this cutoff.
Causal decoder-only models such as TimesFM \citep{das2024timesfm}, Toto \citep{cohen2025toto}, and Timer \citep{liu2024timer} map the representation at each position to a forecast of the following patch (\cref{fig:background:causal_ar_train}).
Trained on fully observed sequences, a single forward pass thus provides a one-patch-ahead forecast from every position of the sequence, reusing the shared history instead of re-encoding it for each forecast.
Patch-wise \ac{ar} inference extends the horizon by feeding decoded predictions back as inputs before predicting the next patch.
\Ac{ar} forecasting can estimate uncertainty over longer horizons by sampling multiple trajectories \citep{salinas2020deepar}.
However, sampling multiple trajectories adds significant computational cost, and feeding predictions back into the model can compound errors \citep{NIPS2015_e995f98d, wen2017multi}.

\begin{figure}[tb]
    \centering
    \begin{subfigure}[b]{0.49\linewidth}
        \centering
        \includegraphics[scale=\modelfigurescale]{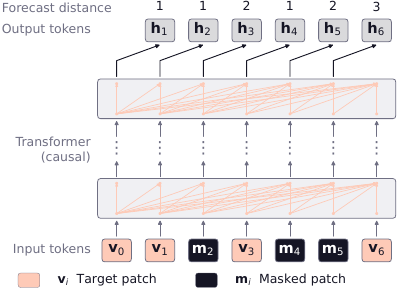}
        \caption{Training.}
        \label{fig:background:causal_dec_cpm_train}
    \end{subfigure}
    \hfill
    \begin{subfigure}[b]{0.49\linewidth}
        \centering
        \includegraphics[scale=\modelfigurescale]{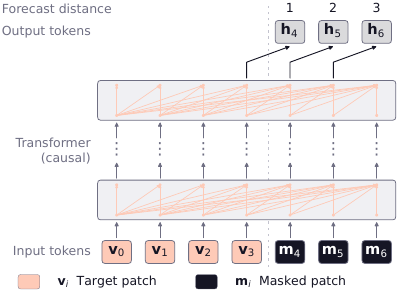}
        \caption{Forecasting.}
        \label{fig:background:causal_dec_cpm_forecast}
    \end{subfigure}
    \caption{Causal decoder-only model with \ac{cpm}: training and forecasting, using the token-level convention of \cref{fig:background:bidir_enc_vs_causal_dec_train}.
    Each output $\finalpatchemb_j$ is computed at position $j-1$ and passed to the prediction head for patch $j$.
    (a) Masking patch $2$ makes $\finalpatchemb_3$ serve a two-patch-ahead forecast from the context ending at patch $1$.
    After visible patch $3$, masked patches $4$ and $5$ allow forecasts for patches $4,5,6$ at distances $1,2,3$.
    Their observed context contains patches $0,1,3$, but not masked patch $2$.
    (b) Context tokens $\patchemb_0,\ldots,\patchemb_3$ followed by $\maskedpatchemb_4,\ldots,\maskedpatchemb_6$ yield the three required outputs $\finalpatchemb_4,\finalpatchemb_5,\finalpatchemb_6$ from positions $3,4,5$.}
    \label{fig:background:causal_dec_cpm}
\end{figure}

\paragraph{\Acl*{cpm}.} \ac{cpm} trains models to forecast several future patches without feeding intermediate predictions back as inputs.
First introduced in TiRex \citep{auer2025tirex} and later adapted to Transformers in \totoTwo \citep{khwaja2026toto2}, \ac{cpm} masks consecutive input patches during training, teaching the model to forecast several patches ahead from the preceding context.
A masked patch has all its values hidden from the model but remains in the input sequence as a token with embedding $\maskedpatchemb_i$.
A run of $r$ masked patches following a visible patch therefore allows forecasts up to $r+1$ patches ahead (\cref{fig:background:causal_dec_cpm_train}).
At inference, the input sequence is constructed from the observed context followed by a block of masked patches covering the forecast horizon (\cref{fig:background:causal_dec_cpm_forecast}).
The model then forecasts the whole block in one forward pass, without feeding intermediate predictions back as inputs.
During training, however, patches masked to create longer-horizon forecasts also remain unavailable as observations for forecasts starting later in the sequence.
The masking pattern therefore determines both the forecast horizons encountered during training and the observations available as context.
In contrast, \ac{dhf}, introduced in \cref{sec:timestral:dhf}, separates horizon sampling from context masking while retaining shared causal context computation.

\section{Timestral}

\begin{figure}[h!]
    \centering
    \includegraphics[scale=\modelfigurescale]{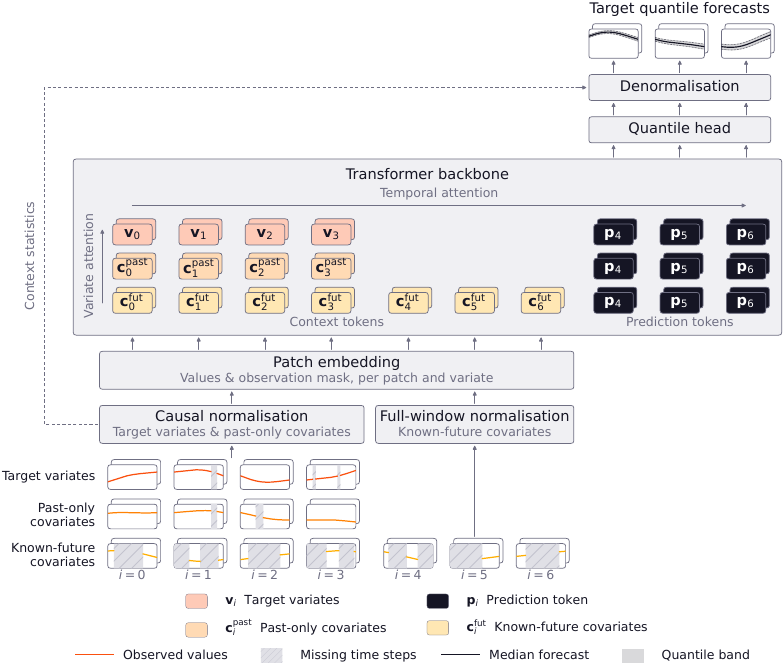}
    \caption{Timestral overview: target variates, past-only covariates, and known-future covariates.
    Input boxes show target values $\mathbf{\targetvalue}_t$ and covariate values $\mathbf{\covariatevalue}_t$ in patches of $P$ consecutive time steps.
    Within the Transformer, context and prediction tokens form two separate streams (\cref{sec:timestral:dhf}): $\patchemb_i$, $\pastcovemb_i$, and $\knowncovemb_i$ denote the context tokens of patch $i$ of a target variate, a past-only covariate, and a known-future covariate, respectively, and $\predtoken_i$ denotes the prediction token at the temporal position of patch $i$.
    Every variate has a prediction token for each forecast patch, but only known-future covariates also have context tokens in the forecast horizon.
    Only the outputs at the prediction tokens of target variates are passed to the quantile head.
    Stacked boxes represent multiple variates.}
    \label{fig:timestral:overview}
\end{figure}

\timestral is a causal encoder-only model for multivariate forecasting that supports both past-only and known-future covariates (\cref{fig:timestral:overview}).
As in bidirectional encoder-only models such as \chronosTwo and \falconTwo, forecasts are produced at dedicated prediction tokens rather than from context representations; unlike them, the context is encoded causally.
Two modelling choices define the dependency structure of \timestral, along time and across variates.
Along time, \ac{dhf} (\cref{sec:timestral:dhf}) decouples the context from the forecast horizon: prediction tokens attend to the context without affecting it.
Across variates (\cref{sec:timestral:covariates}), targets exchange information with each other and receive information from covariates, but never pass information to them; only known-future covariates contribute values from later patches, including the forecast horizon.
\Cref{sec:timestral:backbone} describes the patch embedding, normalisation, Transformer backbone, and quantile head.

\subsection{Temporal Modelling via \acl*{dhf}}
\label{sec:timestral:dhf}
\begin{figure}[htbp]
    \centering
    \includegraphics[scale=\modelfigurescale]{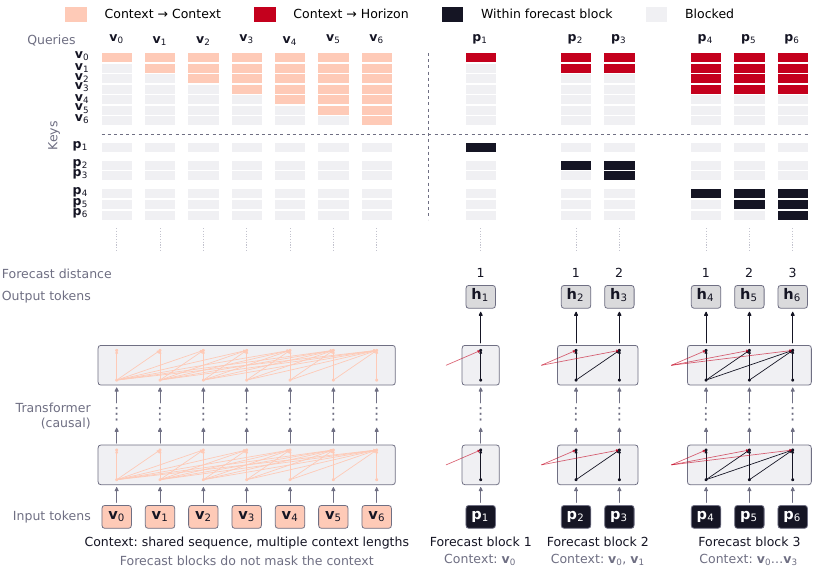}
    \caption{\timestral training with \ac{dhf}.
    Shared context and independent forecast blocks, with the corresponding attention mask above.
    Columns index queries and rows index keys.
    Context tokens attend causally to the context and never to prediction tokens; prediction tokens attend only to the context before their forecast block and to preceding and current prediction tokens in the same block.
    For readability, the red context-to-horizon arrows are drawn short: each stands for a prediction token's attention to all context tokens before its block, as given by the red cells in the mask above.
    Output tokens $\finalpatchemb_j$ are passed to the prediction head; numbers above them give forecast distances in patches.
    The blocks match the forecast distances of \cref{fig:background:causal_dec_cpm_train}, but no context patch is masked to create them.
    Missing-data augmentation of the context is applied independently of the forecast block lengths.}
    \label{fig:timestral:DHP_attention_mask_view}
\end{figure}

\Ac{dhf} uses separate tokens for observed context patches and for forecast positions.
For simplicity, we describe \ac{dhf} for a single target variate.
Temporal attention operates on each variate separately, and all variates of a series share the same forecast blocks.
The context consists of one token $\patchemb_i$ per patch $i$, processed with causal attention, so the representation of patch $i$ depends only on patches $0,\ldots,i$.
As shown in \cref{fig:timestral:DHP_attention_mask_view}, a forecast block covering patches $b,\ldots,b+k-1$ adds prediction tokens $\predtoken_b,\ldots,\predtoken_{b+k-1}$.
These are copies of a single learned embedding, distinguished by their temporal position: $\predtoken_j$ takes the same temporal position as the context token $\patchemb_j$.
Both kinds of tokens are processed by the same Transformer, with shared parameters.
Each $\predtoken_j$ attends to the context tokens before its block and to the prediction tokens $\predtoken_b,\ldots,\predtoken_j$ of its own block, whereas context tokens never attend to prediction tokens.
Context and prediction tokens thus form two separate streams.
The resulting output $\finalpatchemb_j$ is passed to the prediction head to forecast patch $j$ and therefore depends only on $\patchemb_0,\ldots,\patchemb_{b-1}$ and $\predtoken_b,\ldots,\predtoken_j$.
Outputs at context tokens are not passed to the prediction head and receive no direct loss.

To train forecasts at many forecast distances, each training sequence is tiled into consecutive forecast blocks whose lengths are sampled from a block-length distribution.
A block of length $k$ contains one forecast at each distance $1,\ldots,k$, so this distribution determines how often each forecast distance is trained.
Because context tokens never attend to prediction tokens, the context representations are computed once and shared by all blocks of a training sequence.
As in causal decoder-only models (\cref{fig:background:causal_ar_train}), a single training sequence thus provides training signal from forecasts at many starting points, but in \ac{dhf}, the forecast from each starting point can extend several patches ahead.

Unlike \ac{cpm}, \ac{dhf} does not mask context patches to create longer forecast distances.
In \ac{cpm}, a forecast at distance $d$ requires masking the $d-1$ patches before its target, and these patches are then also missing from the context of all later forecasts, so training longer forecast distances removes context.
In \ac{dhf}, each patch in a forecast block has two tokens during training: its prediction token, at which it is forecast, and its context token, which keeps its observed values available to later blocks.
In \cref{fig:timestral:DHP_attention_mask_view}, for instance, the forecasts of the third block depend on $\patchemb_2$, which \ac{cpm} masks.

This decoupling has the following advantages.
First, how often each forecast distance is trained can be tuned without side effects on the context.
Second, missing data can be simulated independently of the forecast distances: optional missing-data augmentation hides variable-length gaps at arbitrary time steps rather than whole patches, which is closer to realistic missing-data distributions, and the hidden values are still scored as targets.
Third, missing values, whether in the data or introduced by augmentation, are flagged in their context tokens and thus represented differently from forecast positions, whereas \ac{cpm} represents a patch masked for forecasting in the same way as a fully missing patch.
These advantages come at the cost of roughly doubling the number of tokens during training, since forecast blocks tile the full sequence so that every patch except the first serves as a forecast target.

At inference, a single forecast block follows the context, so forecasts up to the maximum block length of 32 patches (1,024 time steps) are produced in a single forward pass.
Longer horizons can be forecast block by block, conditioning each block on the predicted medians of the previous ones, as in \totoTwo \citep{khwaja2026toto2}.

\subsection{Multivariate and Covariate Modelling}
\label{sec:timestral:covariates}
Each variate of a multivariate series is represented by its own sequence of tokens and has one of three roles: target variate, past-only covariate, or known-future covariate (\cref{fig:timestral:overview}).
Every variate has context and prediction tokens, but only the outputs at prediction tokens of target variates are used as forecasts.
Roles are not embedded in the tokens and enter the model only through the attention masks and the choice of normalisation (\cref{sec:timestral:backbone}).
Along time, target variates and past-only covariates are processed causally, as described in \cref{sec:timestral:dhf}.
The context tokens of known-future covariates instead attend over the entire window, including the forecast horizon, where they hold the known future values.
Normalisation follows the same distinction (\cref{sec:timestral:backbone}).

Similar to \chronosTwo and \totoTwo \citep{ansari2025chronos2, khwaja2026toto2}, variate attention exchanges information between the variates of a series at the same patch position, among context tokens and among prediction tokens.
It uses no positional encoding, so the order of the variates does not affect the forecasts.
To prevent future information leakage, known-future covariates attend only to each other.
Otherwise, a known-future covariate could read the value of a target variate or past-only covariate at patch $i$ and, through its temporal attention, carry it back to earlier patches, from which it would reach the forecast of patch $i$.
Past-only covariates do not attend to target variates either; this is a modelling choice that treats covariates as exogenous, so that their representations do not depend on target values.
Forecasts can thus use future values of known-future covariates, but never future values of target variates or past-only covariates.

During training, some variates in a random subset of multivariate series are assigned the role of past-only or known-future covariates, while the others remain target variates.
These covariates are inputs only and are not scored, so ordinary multivariate series become forecasting tasks with covariates.

\subsection{Architecture}
\label{sec:timestral:backbone}
Each patch of $P = 32$ time steps of a variate is embedded into a single token by a two-layer residual \ac{mlp}.
Its input consists of the $P$ normalised values and $P$ binary indicators of missing time steps, at which the values are set to zero, so that missing values are distinguished from observed zeros.
Similar to previous work, values are standardised causally with the mean and standard deviation of all observed values of their variate up to the end of the current patch \citep{cohen2025toto}, and an inverse hyperbolic sine is applied to the standardised values \citep{ansari2025chronos2}.
Known-future covariates, whose values are known over the entire window, are instead normalised with statistics over this window, and binary ones, such as holiday flags, are left unchanged.

The backbone is a pre-norm Transformer with RMSNorm \citep{zhang2019rmsnorm} and SwiGLU feed-forward layers \citep{shazeer2020glu}.
Each block applies temporal attention and a feed-forward layer, and every fourth block additionally applies variate attention between the two.
Both attention types use gated attention with an elementwise sigmoid output gate \citep{qiu2025gated}.
Temporal attention encodes token positions with rotary position embeddings \citep{su2024roformer} on half of each head's dimensions.
No timestamps, frequency information, or role or position embeddings are added; positions enter only through the rotary embeddings of temporal attention.
Neither dropout nor stochastic depth is used.

After a final RMSNorm, a two-layer residual \ac{mlp} maps each output $\finalpatchemb_j$ to 21 quantiles for each of the $P$ time steps of patch $j$.
The head predicts the median directly and the other quantiles by adding or subtracting cumulative sums of positive (softplus) increments, preventing quantile crossing by construction.
Each forecast patch is standardised with the statistics of all values up to the end of the preceding patch, and its predicted quantiles are mapped back to the original scale by inverting the asinh and this standardisation.
Within a forecast block, these values include the earlier patches of the block, which are unknown at inference and are therefore replaced by their predicted medians.
The same statistics are used during training to standardise the targets, so that training and inference match.
\Cref{tab:timestral:sizes} lists the model sizes.

\begin{table}[tb]
    \centering
    \small
    \caption{Timestral model sizes.
    All models use a head dimension of 64, sigmoid-gated attention, and variate attention in every fourth block.}
    \label{tab:timestral:sizes}
    \begin{tabular}{@{}lrrrrr@{}}
        \toprule
        Model & $d_{\mathrm{model}}$ & Blocks & Heads & SwiGLU width & Parameters \\
        \midrule
        \timestralSize{5M}   & 192 & 8  & 3  & 512  & 5.2M \\
        \timestralSize{28M}  & 384 & 12 & 6  & 1024 & 27.8M \\
        \timestralSize{123M} & 640 & 20 & 10 & 1712 & 122.6M \\
        \timestralSize{209M} & 768 & 24 & 12 & 2048 & 209.3M \\
        \bottomrule
    \end{tabular}
\end{table}

\section{Results}
\label{sec:results}

We describe the data and training setup shared by the four \timestral models, then report their forecasting performance on public benchmarks.

\subsection{Training Setup}

All four \timestral models are pretrained exclusively on synthetic data with a context length of
8,192 time steps and then post-trained on a mixture of real and synthetic data with a context
length of 32,768 time steps for substantially fewer steps.

\paragraph{Training data.}
\label{sec:timestral:synthdata}
The synthetic corpus includes both univariate signals and multivariate examples, covering temporal dynamics and cross-series dependencies.
We generate univariate series with diverse trends, seasonal patterns, and noise characteristics across temporal resolutions.
Additional patterns include regime changes, localised events, level shifts, and count-valued observations.
We construct multivariate examples by introducing linear and nonlinear dependencies between base univariate signals, using structured synthetic priors~\citep{ansari2025chronos2, xie2026cauker, lenaour2026tsicl, podest2026tirex2}.
Selected variates serve as past-only or known-future covariates, providing training examples with different input availability patterns.

\paragraph{Training objective.}
We train with the pinball loss~\citep{koenker1978regression}, without auxiliary objectives.
For a target value $y$ and a predicted quantile $\hat{q}_\tau$ at level $\tau$, the loss is
\begin{equation}
\rho_\tau(e) = \max\!\left(\tau e,\, (\tau - 1)e\right),
\qquad e = y - \hat{q}_\tau.
\end{equation}
Minimising this loss trains $\hat{q}_\tau$ to estimate the conditional
$\tau$-quantile of $y$~\citep{koenker1978regression}.
The prediction head outputs 21 quantile levels,
$\mathcal{T} = \{0.01, 0.05, 0.10, \dots, 0.95, 0.99\}$.
We average the loss equally across quantile levels and scored positions:
\begin{equation}
\mathcal{L} = \frac{1}{|\mathcal{T}|\sum_i w_i}
\sum_i w_i \sum_{\tau \in \mathcal{T}}
\rho_\tau\!\left(y_i - \hat{q}_{\tau,i}\right),
\end{equation}
where $i$ indexes target values within forecast blocks and $w_i$ is one for scored positions and zero otherwise.
Only observed target values within a forecast horizon are scored.
Context representations and covariates receive no direct loss; they are trained through their influence on the scored predictions.
Targets and predictions are compared in the model's normalised space, reducing differences in loss scale between series.

\subsection{Forecasting Performance}

We evaluate \timestral on GIFT-Eval~\citep{aksu2024giftplus}, TIME~\citep{qiao2026time}, and fev-bench~\citep{shchur2025fevbench}.
On fev-bench, we report skill scores for \ac{mase} and \ac{sql}.
On GIFT-Eval, \timestral uses past-only covariates on the few datasets that provide them.
On fev-bench, it uses the available past-only and known-future covariates according to each task's input specification.

\begin{figure}[tb]
    \centering
    \includegraphics[width=1.0\linewidth,page=1]{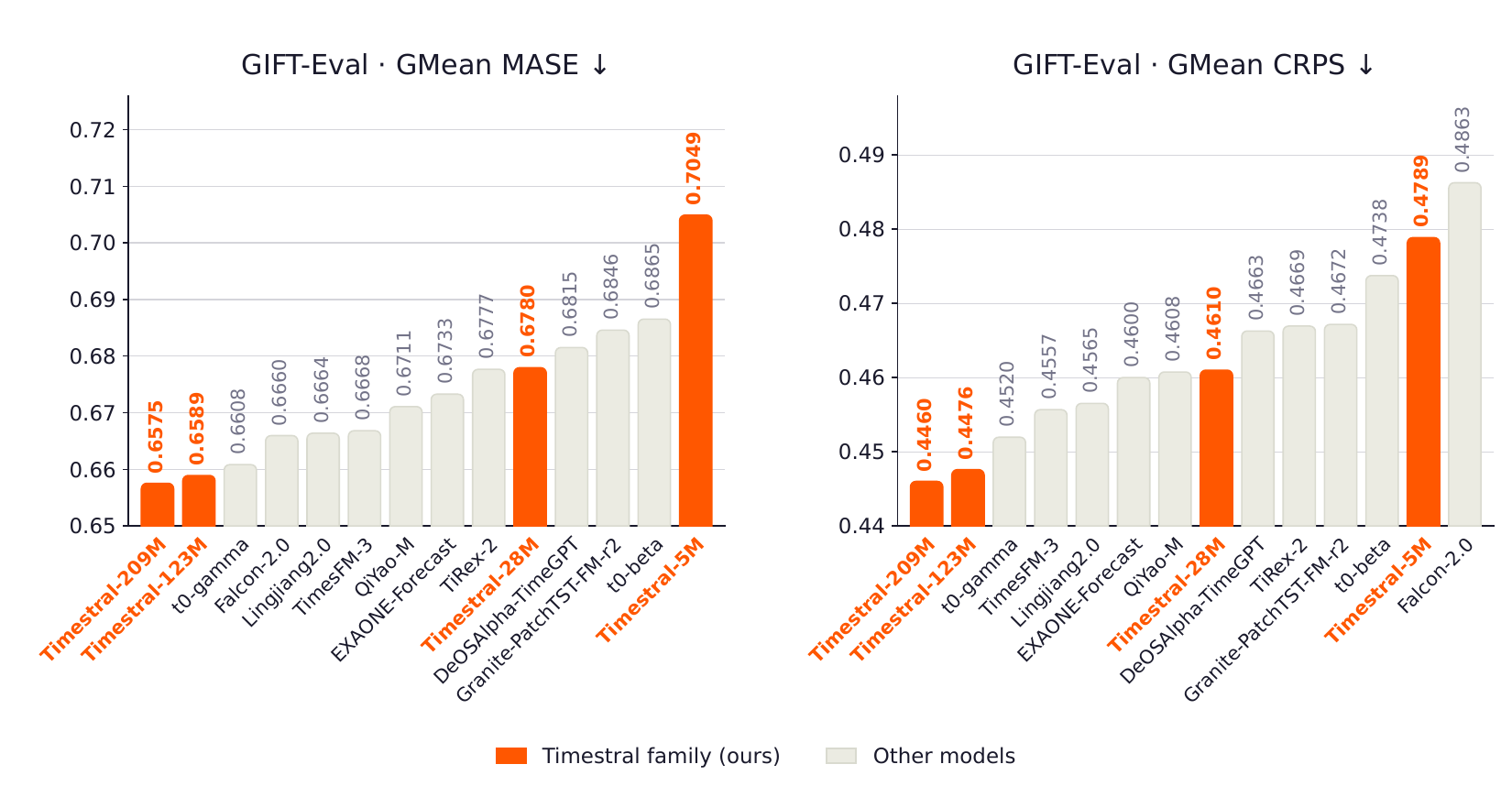}
    \caption{GIFT-Eval results for all four \timestral models and the ten leading competitors by aggregate \ac{mase}. Both panels show the same models, ranked separately by each metric. Lower is better.
    }
    \label{fig:results:gift}
\end{figure}

\begin{figure}[tb]
    \centering
    \includegraphics[width=1.0\linewidth,page=2]{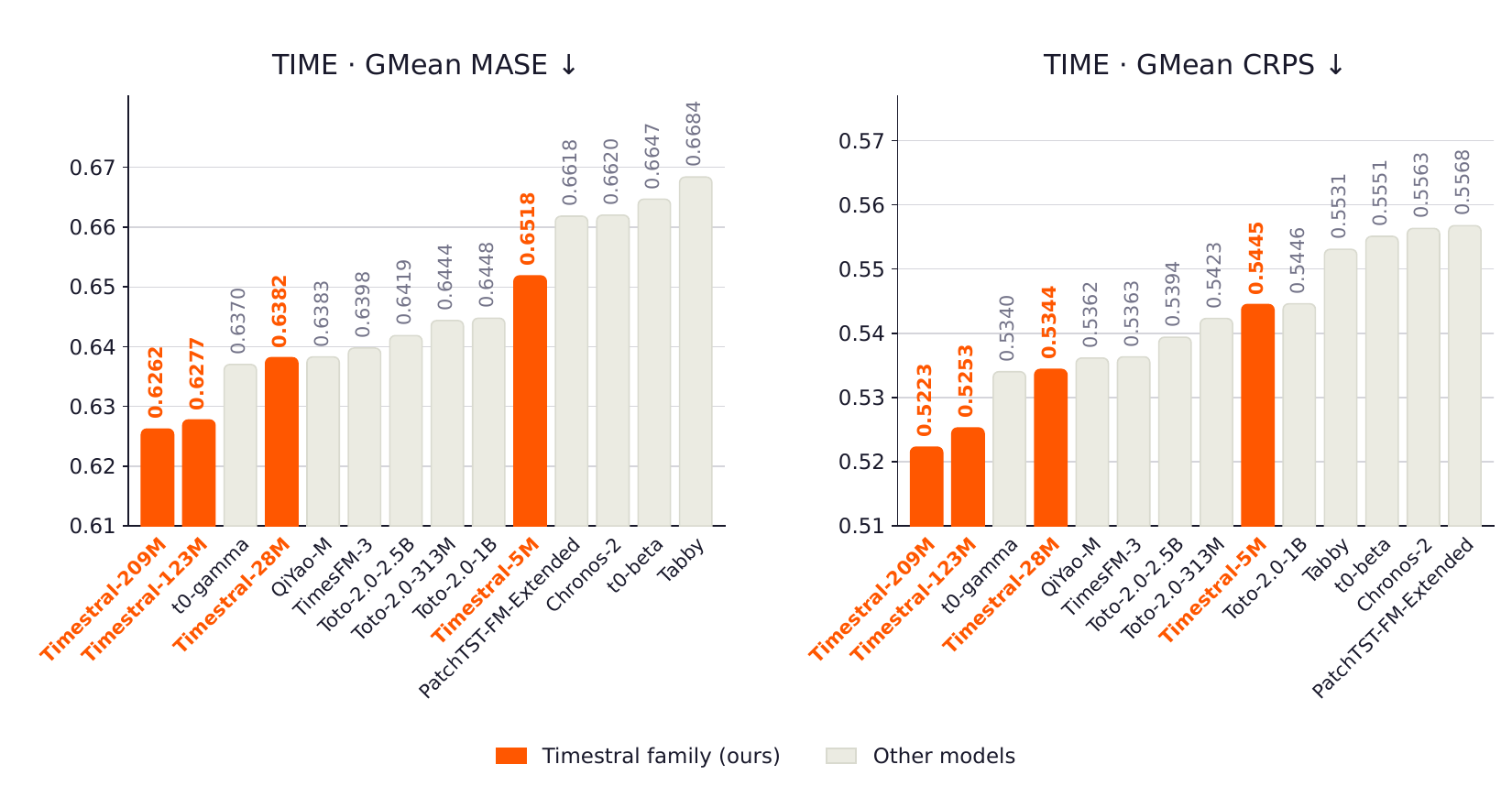}
    \caption{TIME results for all four \timestral models and the ten leading competitors by aggregate \ac{mase}. Both panels show the same models, ranked separately by each metric. Lower is better on both metrics.}
    \label{fig:results:time}
\end{figure}

\begin{figure}[tb]
    \centering
    \includegraphics[width=1.0\linewidth]{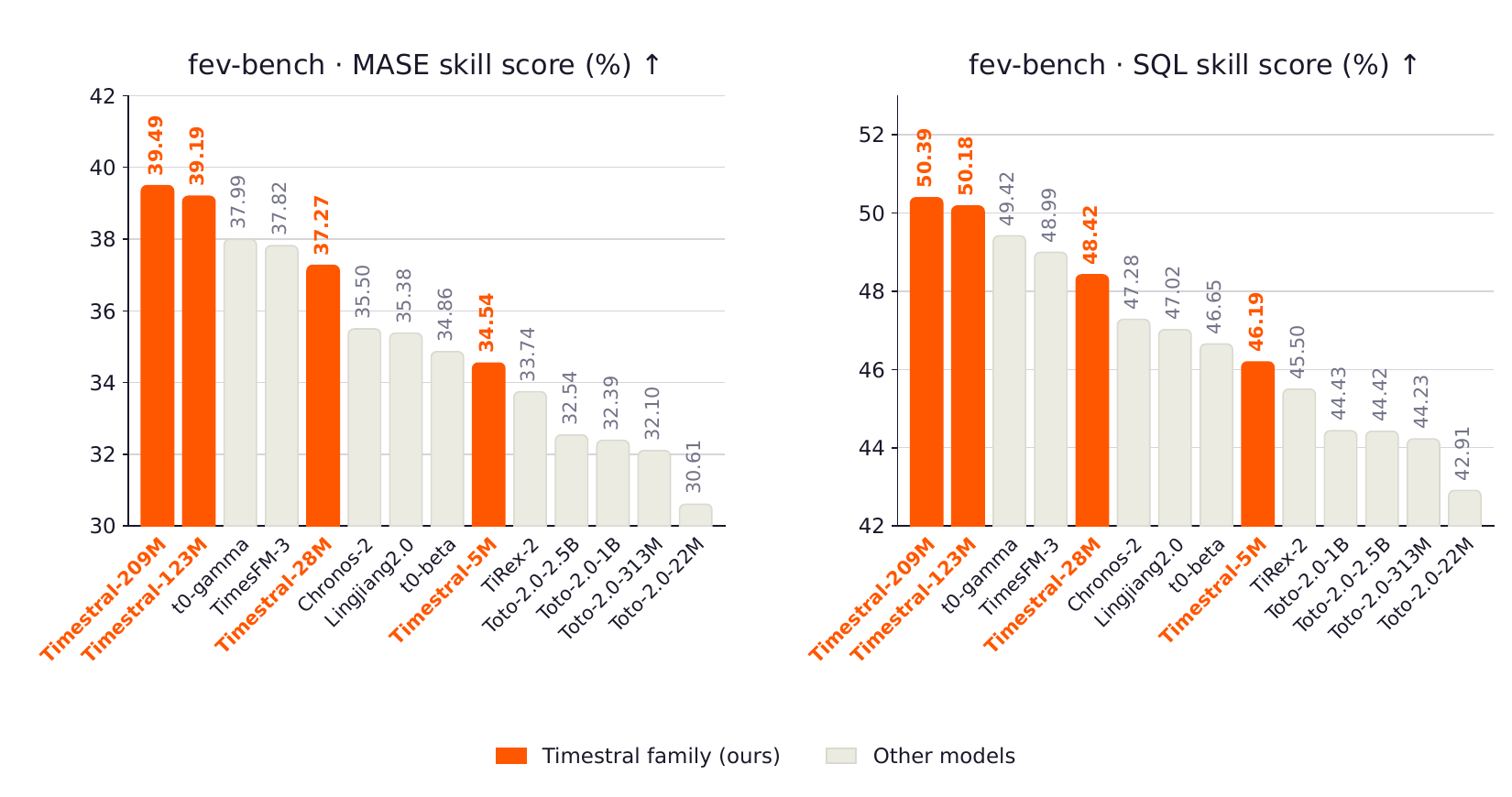}
    \caption{fev-bench results for all four \timestral models and the ten leading competitors by MASE skill score. Both panels show the same models, ranked separately by each metric. Higher is better for both MASE and \acs{sql} skill scores.}
    \label{fig:results:fev}
\end{figure}

\paragraph{\timestral sets a new state of the art across GIFT-Eval, TIME, and fev-bench.}
\Cref{fig:results:gift,fig:results:time,fig:results:fev} compare all four \timestral sizes with other leading entries on each benchmark, and \cref{tab:benchmark-scores} in \cref{app:detailed-results} reports the scores of all compared models.
\timestralSize{209M} ranks first on all six metrics among the models compared: it has the lowest \ac{mase} and \ac{crps} on GIFT-Eval and TIME and the highest \ac{mase} and \ac{sql} skill scores on fev-bench, using a single checkpoint.
Pairwise win rates in \cref{app:win-rates} complement these aggregate scores by measuring how often \timestral improves over each competitor.
\Cref{app:ind-eval} additionally reports results on the industrial benchmark IND-Eval~\citep{haugland2026indevalaligningevaluationsimulation}, where \timestralSize{209M}, \timestralSize{123M}, and \timestralSize{28M} take the top three places.

\paragraph{Every \timestral size lies on the Pareto front.}
\Cref{fig:pareto} plots accuracy against parameter count: all four \timestral models lie on the Pareto front, and the smaller ones outperform much larger models.
\timestralSize{123M} ranks second on all six metrics, ahead of every model outside the family, and both \timestralSize{123M} and \timestralSize{209M} have fewer parameters than \tzeroGamma (256M)~\citep{theforecastingcompany2026t0gamma} and \timesfmThree (331M).
\timestralSize{28M} outperforms \chronosTwo and \totoTwoSize{2.5B}, about 4 and 90 times larger, on all six metrics, and on fev-bench \timestralSize{5M} outperforms every \totoTwo model, including \totoTwoSize{2.5B} with about 480 times as many parameters.

\subsection{Inference Speed}
\label{sec:results:inference}

\timestral does not require a GPU at inference.
A forecast of up to 1,024 time steps is produced in a single forward pass (\cref{sec:timestral:dhf}), so its cost depends mainly on the context length and the number of variates.
\Cref{tab:inference} reports the latency and throughput of each model for a representative workload: forecasting one day ahead (96 steps) of a univariate series without covariates, from about three weeks of 15-minute history (2,048 steps).

\paragraph{Setup.}
We time the \texttt{forecast} method of the released inference package with the released checkpoints, end to end, with bfloat16 weights and \texttt{torch.compile}; the bfloat16 forecasts differ from float32 ones by less than 0.2\%.
Each forecast uses symmetric averaging: the model also forecasts the negated series and averages the reflected forecast with the original one, which requires two forward passes.
CPU numbers are measured on a 48-core server CPU (AMD EPYC 9454) and GPU numbers on one NVIDIA H100, with PyTorch 2.14; we report the median over repeated runs after warm-up.

\begin{table}[tb]
    \centering
    \small
    \caption{Inference latency and throughput for forecasting 96 steps of a univariate series without covariates from a context of 2,048 steps, with symmetric averaging. CPU latency is the median of 50 runs at batch size 1. CPU throughput is per core at batch size 16, and GPU throughput is on one NVIDIA H100 at batch size 256. Checkpoint sizes are for float32 weights.}
    \label{tab:inference}
    \begin{tabular}{@{}lrrrrrr@{}}
        \toprule
        & & & \multicolumn{3}{c}{CPU} & GPU \\
        \cmidrule(lr){4-6} \cmidrule(l){7-7}
        Model & Parameters & Checkpoint size & 8 threads & 1 thread & series/s & series/s \\
        \midrule
        \timestralSize{5M}   & 5.2M & 21\,MB & 17\,ms & 16\,ms & 106 & 18{,}770 \\
        \timestralSize{28M}  & 27.8M & 111\,MB & 31\,ms & 56\,ms & 24 & 10{,}730 \\
        \timestralSize{123M} & 122.6M & 490\,MB & 85\,ms & 210\,ms & 5.9 & 4{,}990 \\
        \timestralSize{209M} & 209.3M & 837\,MB & 113\,ms & 345\,ms & 3.5 & 3{,}650 \\
        \bottomrule
    \end{tabular}
\end{table}

\paragraph{A forecast takes 17 to 113\,ms on 8 CPU threads.}
Latency grows with model size, and additional threads mainly benefit the larger models: from 1 to 8 threads, \timestralSize{209M} becomes 3.1 times faster, whereas \timestralSize{5M} is already as fast on a single core.

\paragraph{Smaller models trade some accuracy for speed.}
\timestralSize{28M} is 3.6 times faster than \timestralSize{209M}, and \timestralSize{123M} has about 0.2\% higher GIFT-Eval \ac{mase} than \timestralSize{209M} at 75\% of its latency (\cref{fig:pareto,tab:inference}).

\paragraph{Throughput.}
One CPU core forecasts 3.5 (\timestralSize{209M}) to 106 (\timestralSize{5M}) series per second, so forecasting 100,000 series every 15 minutes requires 2 to 32 cores.
One NVIDIA H100 forecasts 3{,}650 to 18{,}770 series per second, that is, 100,000 series in 5 to 27 seconds.

\begin{figure}[tb]
    \centering
    \includegraphics[width=1.0\linewidth]{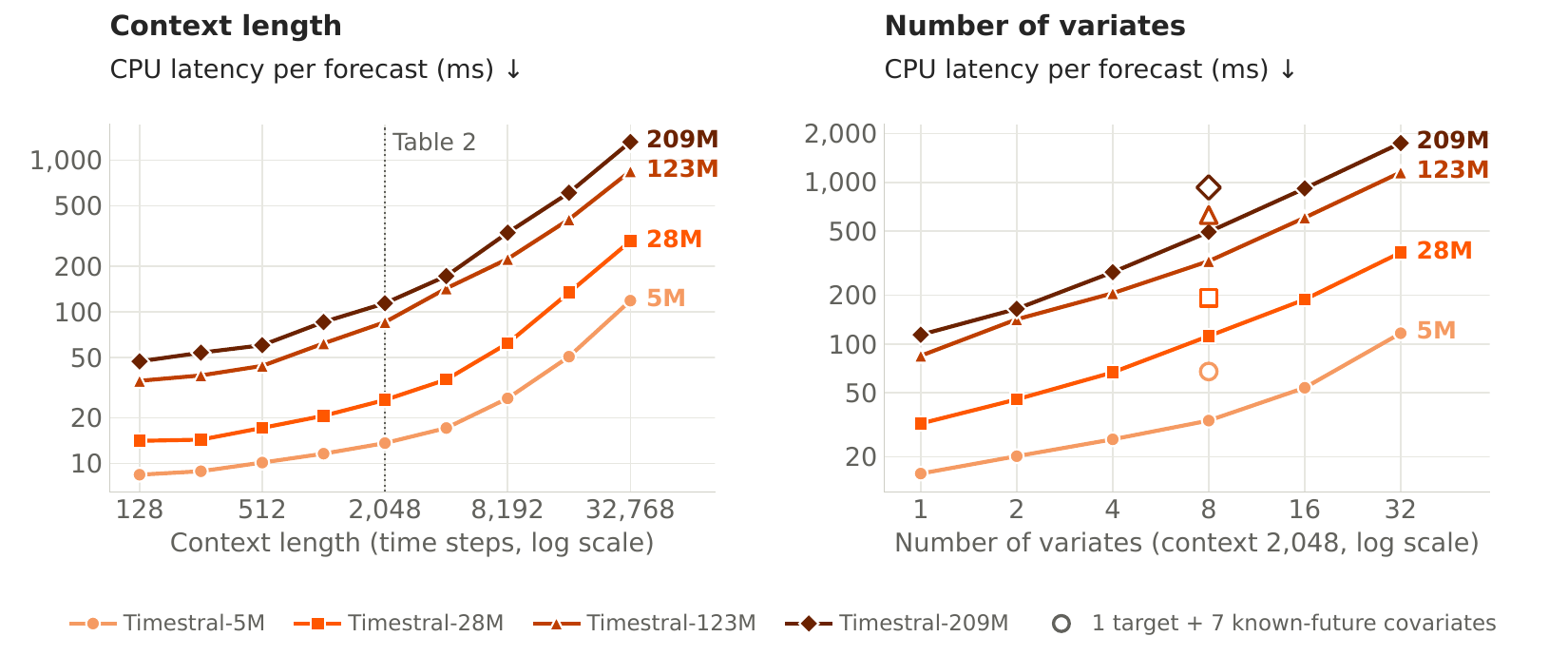}
    \caption{CPU latency per forecast against context length (left, one variate) and number of target variates (right, context of 2,048 steps), on 8 threads with the settings of \cref{tab:inference} (median of 20 runs). Hollow markers: one target with seven known-future covariates. Both axes are logarithmic.}
    \label{fig:inference-scaling}
\end{figure}

\paragraph{Context length and number of variates.}
\Cref{fig:inference-scaling} varies the context length and the number of variates.
Up to 2,048 steps, latency changes little: it roughly doubles from 128 to 2,048 steps.
Beyond, it grows nearly in proportion to the context, by about 9 to 12 times from 2,048 to 32,768 steps, the longest context used in training; at that length, a forecast takes 119\,ms with \timestralSize{5M} and 1.3\,s with \timestralSize{209M}.
A series with 32 variates takes 7 to 15 times as long as a univariate series.
Covariates cost more than target variates: one target with seven known-future covariates takes 1.7 to 2.0 times as long as eight target variates.

\section{Conclusion}
\label{sec:conclusion}
We introduced \timestral, a family of four foundation models with 5M to 209M parameters for probabilistic univariate, multivariate, and covariate-informed forecasting.
\Ac{dhf} separates prediction tokens from the causally encoded context, so that observed values serve both as forecast targets and as context for later forecasts, and longer forecast distances are trained without masking context.
Role-aware attention masks let prediction tokens use known-future covariates at every forecast origin, while forecasts never use future values of target variates or past-only covariates.
The models are pretrained exclusively on synthetic data and then post-trained on a mixture of real and synthetic data with extended context.

\timestralSize{209M} ranks first among the models compared on every reported metric of GIFT-Eval, TIME, and fev-bench using a single checkpoint.
It leads by a wider margin on IND-Eval.
\timestralSize{123M} ranks second, ahead of every model outside the \timestral family, and all four sizes lie on the Pareto front of accuracy against size.
Accuracy still improves from 123M to 209M parameters, which motivates training larger models.
Beyond standalone forecasting, \timestral can be embedded in existing monitoring pipelines or combined with language models in agentic workflows.
A further direction is to use the context representations as embeddings for tasks such as classification and anomaly detection.

\section*{Core contributors}
Adrien Petralia, Alexandre Gavaudan, Artjom Joosen, Etienne Goffinet, Richard Kurle, Yassine El Ouahidi, Yassir Bendou.

\section*{Contributors}

Abdelaziz Bounhar, Abhijeet Somani, Aditi Kabra, Adrian Valente, Adrien Sadé, Ahmed Djermani, Alan
Jeffares, Albert Jiang, Alberto Bietti, Aleksandr Timashov, Alexandre Cahill, Alexandre Laval, Alexandre
Sablayrolles, Alexandros Sanchez Vassopoulos, Amélie Héliou, Amos You, André Jonasson, Andreas
Fürst, Andrew Bai, Andrew Ehrenberg, Andrew Zhao, Andy Lo, Angele Lenglemetz, Anmol Agarwal, Antonia Calvi,
Arata Suzuki, Arjun Majumdar, Arthur Fournier, Ashwani Kumar, Augustin Lombard, Avinash
Sooriyarachchi, Aylin Guliz Akkus, Aysenur Karaduman, Baptiste Bout, Baptiste Rozière, Barna
Pásztor, Baudouin De Monicault, Benjamin Holzschuh, Benjamin Tibi, Benno Stein, Bernhard Stadlbauer,
Bhargavi Bhaskar, Binbing Hou, Błażej Osiński, Boris van Breugel, Camille Le Scao, Chaoran Yu,
Charlotte Cronjäger, Chen-Yo Sun, Chris Bamford, Christian Wallenwein, Christophe Renaudin, Clémence Lanfranchi, Colin Bonatti, Corentin Barreau, Corentin Sautier,
Cristiana-Diana Diaconu, Cyprien Courtot, Daniel Lopes, Daniel Marczak, Darius Dabert, Diego de las
Casas, Dominik Nuss, Dylan Rubini, Dzmitry Soupel, Elizaveta Demyanenko, Elliot Chane-Sane, Emilien
Fugier, Emmanuel Gottlob, Erik Aas, Etienne Duchesne, Fabian Paischer, Fabian Schlager, Faruk Ahmed,
Federico Baldassarre, Félix Rosseeuw, Filip Szatkowski, Filippo Vicentini, Florian Wiesner,
Gabrielle Berrada, Gaëtan Ecrepont, Gaspard Blanchet, Gaspard Donada-Vidal, Gaurav Mahajan, Gauthier
Guinet, Genevieve Hayes, Georgii Novikov, Georgios Christoglou, Giada Pistilli, Gonzague de
Carpentier, Graeme Nail, Guillaume Breton, Guillaume Kunsch, Guillaume Lample, Guillaume Martin,
Guillaume Raille, Guillaume Richard, Guillem Simeon, Gunjan Dhanuka, Gunshi Gupta, Hamza Senhaji
Rhazi, Han Zhou, Hédi Hadiji, Himanshu Jaju, Hope McGovern, Hubert de Parseval, Hugo Cisneros, Hugo
Germain, Hugo Thimonier, Hugo Venturini, Hyeonju Lee, Hyunwoo Oh, Iliass Lasri, Ilyas Oulkadda,
Jacques Sun, Jan Ludziejewski, Jason Rute, Jean de Becdelièvre, Jean Quentin, Jean-Malo Delignon,
Jessica Chudnovsky, Jianchao Li, Jie Zhang, Jin Bin Ou, Joachim Studnia, Joep Barmentlo, Johannes
Brandstetter, John Harvill, Jonas Amar, Jonas Schweizer, Jonathan Roberts, Joséphine Delas, Josselin Somerville, Jules Depersin, Julien Denize, Julien Tauran, Julien Verlaguet, Kartik
Khandelwal, Khyathi Raghavi Chandu, Kilian Tep, Konstantin Mishchenko, Kris Szczepaniak, Kush Jain,
Larissa Laich, Laura Calem, Laurence Aitchison, Laurent Callot, Laurent Fainsin, Léo Cotteleer,
Léonard Blier, Loïc Chadoutaud, Loïck Chambon, Lorenzo Proietti, Louis Faucon, Louis Martin, Louis
Serrano, Lu Li, Luc Boudier, Lucas Fernandes Martins, Lucas Mebille, Lucile Saulnier, Ludovic Ho
Fuh, Luis Montero, Łukasz Borchmann, Maarten Buyl, Malik Hassan Raza, Manon Chossegros,
Marcin Możejko, Margaret Jennings, Markus Hennerbichler, Martin Alexandre, Mathieu Poirée,
Mathieu Schmitt, Mathilde Guillaumin, Mathis Felardos, Matthieu André, Matthieu Dinot, Matthieu
Futeral, Maurits Bleeker, Mauro Comi, Max Mynter, Maxim Berman, Maxime Darrin, Maxime Langelier,
Maxime Louis, Maximilian Augustin, Maximilian Müller, Mehdi Bouaziz, Melina Jingting Laimon, Mert
Unsal, Mia Chiquier, Michael Pilcer, Michał Pietruszka, Michał Zając, Mickaël Seznec, Mikhail
Biriuchinskii, Minh-Quang Pham, Minwoo Kang, Mircea Lica, Morgane Rivière, Namit Katariya,
Natasha Ørregaard Klingenbrunn, Nathan Bout, Nathan Grinsztajn, Nathan Simpson, Neeraj Aggarwal,
Neil Ashton, Neil Mallinar, Nick Bührer, Nicolas Moreau, Nicolò Lucchesi, Ola Mysiak, Oliver Leicht,
Olivier Bousquet, Olivier Koch, Parag Jain, Patricia Wang, Patrick Blies, Patrick von Platen, Patryk
Saffer, Paul Jacob, Paul Teiletche, Paul Wambergue, Paula Kurylowicz, Pavan Kumar Reddy,
Pavel Kuksa, Philippe Formont, Philippe Pinel, Philomène Chagniot, Pierre Glandon, Pierre Stock, Pierre-André
Savalle, Piotr Miłoś, Poojan Shah, Prateek Gupta, Pravesh Agrawal, Pui Ki Chan,
Quentin Desreumaux, Quentin Selle, Quentin Torroba, Quentin Wargnier, Quercus Hernandez, Ram
Ramrakhya, Randall Isenhour, Ranjit Parva, Raphaël Balouka, Raphaël Rialland, Raul Perez Pelaez,
Reinhard Sonnleitner, Remi Delacourt, Rishi Shah, Rob Romijnders, Rohan Shah, Rohin Arora, Romain
Sauvestre, Roman Soletskyi, Ronan Riochet, Rosalie Millner, Rupert Menneer, Sagar Vaze, Samuel
Barry, Samuel Belkadi, Samuel Humeau, Samuele Papa, Sanchit Gandhi, Sandeep Cranganore, Sandeep
Subramanian, Sarthak Mittal, Saskia Adaime, Scott Matthews, Sean Cha, Sebastian Kaltenbach,
Seungjae Jung, Shashwat Verma, Sherif Waly, Shrimai Prabhumoye, Siddhant Waghjale, Siddharth Gandhi,
Simon Lepage, Simon Sorg, Soham Ghosh, Sophie Marbach, Srijan Mishra, Stanislas Lange, Steve Hong, Sumukh Aithal, Szymon
Antoniak, Tarun Kumar Vangani, Teddy Turquet, Teven Le Scao, Théo Cachet, Thibaut
Lavril, Thomas Chabal, Thomas Coste, Thomas Defard, Thomas Foubert,
Thomas Loux, Thomas Robert, Thomas Wang, Tianyu Zhang, Tim Lawson, Timothée Lacroix, Timothy Wallis, Tobias
Kronlachner, Tom Bewley, Tom Edwards, Tomas Hodan, Triantafyllos Afouras, Tuhin Das, Tyler Wang, Udit Saxena, Ulrick Blé,
Umar Jamil, Umberto Tomasini, Valentin Macé, Van Phung, Venkata Reddy, Vibha Masti, Victor Jouault,
Victor Letzelter, Victor Paltz, Victor Poucheret, Vincent Maladière, Vincent Pfister, Virgile
Richard, Vladislav Bataev, Wassim Bouaziz, Wen Ding Li, William Flanagan, William Havard, William
Marshall, Witold Jędrzejewski, Xinghui Li, Xingran Guo, Xinyu Yang, Xuanyu Zhang, Yann Dreze, Yann
Viegas, Yannic Neuhaus, Yassine Turki, Yessin Moakher, Yi He, Yihan Wang, Yimu Pan, Yuying Zhu, Yves Martin des Taillades, Zaccharie Ramzi, Zhenlin Xu, Zsófia Csákány.

\newpage

{
\small
\bibliographystyle{plainnat}
\bibliography{references}
}

\clearpage

\appendix
\crefalias{section}{appendix}
\crefalias{subsection}{appendix}

\section{Detailed Results}
\label{app:detailed-results}

This appendix expands on the results in \cref{sec:results}.
We report aggregate scores for all compared models on GIFT-Eval, TIME, and fev-bench, followed by an evaluation on IND-Eval, which focuses on industrial forecasting tasks.
Pairwise win rates complement these aggregate results by showing how often each \timestral model outperforms the selected competitors.

\subsection{GIFT-Eval, TIME, and fev-bench}
\label{app:benchmark-scores}

\Cref{tab:benchmark-scores} reports aggregate scores for all models included in the benchmark bars or Pareto plot at the date of the release (October 15th 2026). 
The table includes all four released \timestral sizes. \timestralSize{209M} ranks first on all six metrics, and \timestralSize{123M} ranks second, ahead of every model outside the \timestral family in this comparison.

\begin{table*}[!htbp]
\centering
\footnotesize
\setlength{\tabcolsep}{2.5pt}
\renewcommand{\arraystretch}{1.0}
\caption{Aggregate results for leading models on GIFT-Eval, TIME, and fev-bench.
GIFT-Eval and TIME report geometric-mean errors relative to Seasonal Naive (lower is better); fev-bench reports MASE and \acs{sql} skill scores in percent (higher is better). \acs{sql} scales quantile loss by the context's mean absolute seasonal error. 
Bold and underlined scores denote first and second place before rounding; dashes indicate missing scores. 
For \tirexTwo, PT and ZS denote pretrained and zero-shot GIFT-Eval submissions, respectively; its fev-bench entry uses a benchmark-specific checkpoint.}
\label{tab:benchmark-scores}
\begin{tabular}{@{}lrrrrrr@{}}
\toprule
& \multicolumn{2}{c}{GIFT-Eval} & \multicolumn{2}{c}{TIME} & \multicolumn{2}{c}{fev-bench} \\
\cmidrule(lr){2-3}\cmidrule(lr){4-5}\cmidrule(l){6-7}
Model & \acs{mase} $\downarrow$ & \acs{crps} $\downarrow$ & \acs{mase} $\downarrow$ & \acs{crps} $\downarrow$ & \shortstack[r]{\acs{mase} skill\\score (\%) $\uparrow$} & \shortstack[r]{\acs{sql} skill\\score (\%) $\uparrow$} \\
\midrule
\timestralSize{209M} & \textbf{0.6575} & \textbf{0.4460} & \textbf{0.6262} & \textbf{0.5223} & \textbf{39.49} & \textbf{50.39} \\
\timestralSize{123M} & \underline{0.6589} & \underline{0.4476} & \underline{0.6277} & \underline{0.5253} & \underline{39.19} & \underline{50.18} \\
\timestralSize{28M} & 0.6780 & 0.4610 & 0.6382 & 0.5344 & 37.27 & 48.42 \\
\timestralSize{5M} & 0.7049 & 0.4789 & 0.6518 & 0.5445 & 34.54 & 46.19 \\
\midrule
\chronosTwo & 0.6978 & 0.4854 & 0.6620 & 0.5563 & 35.50 & 47.28 \\
Chronos-Bolt & -- & -- & 0.7328 & 0.6200 & 26.52 & 38.89 \\
DeOSAlpha-TimeGPT & 0.6815 & 0.4663 & -- & -- & -- & -- \\
EXAONE-Forecast & 0.6733 & 0.4600 & -- & -- & -- & -- \\
\falconTwo & 0.6660 & 0.4863 & -- & -- & -- & -- \\
Falcon-X & 0.6873 & 0.4857 & -- & -- & -- & -- \\
FlowState-r1 & -- & -- & -- & -- & 27.83 & 39.91 \\
FlowState-r1.1 & 0.7015 & 0.4866 & -- & -- & -- & -- \\
Granite-PatchTST-FM-r2 & 0.6846 & 0.4672 & -- & -- & -- & -- \\
Kairos-50M & 0.7422 & 0.5482 & -- & -- & -- & -- \\
Kairos-23M & 0.7480 & 0.5544 & 0.7395 & 0.6440 & -- & -- \\
Kairos-10M & 0.7527 & 0.5541 & -- & -- & -- & -- \\
Lingjiang2.0 & 0.6664 & 0.4565 & -- & -- & 35.38 & 47.02 \\
Moirai-2.0 & 0.7281 & 0.5164 & 0.7030 & 0.5885 & 27.22 & 39.31 \\
PatchTST-FM-Extended & -- & -- & 0.6618 & 0.5568 & -- & -- \\
QiYao-M & 0.6711 & 0.4608 & 0.6383 & 0.5362 & -- & -- \\
Sundial-base & 0.7502 & 0.5590 & 0.7576 & 0.6634 & 24.75 & 33.42 \\
t0-beta & 0.6865 & 0.4738 & 0.6647 & 0.5551 & 34.86 & 46.65 \\
\tzeroGamma & 0.6608 & 0.4520 & 0.6370 & 0.5340 & 37.99 & 49.42 \\
Tabby & -- & -- & 0.6684 & 0.5531 & -- & -- \\
Timer-S1 & 0.6934 & 0.4853 & 0.7021 & 0.5886 & -- & -- \\
TimesFM-2.5 & 0.7050 & 0.4903 & 0.6686 & 0.5674 & 30.20 & 42.20 \\
\timesfmThree & 0.6668 & 0.4557 & 0.6398 & 0.5363 & 37.82 & 48.99 \\
TiRex & 0.7158 & 0.4885 & 0.6831 & 0.5731 & 30.01 & 42.58 \\
\tirexTwo{} (fev) & -- & -- & -- & -- & 33.74 & 45.50 \\
\tirexTwo{} (PT) & 0.6777 & 0.4669 & -- & -- & -- & -- \\
\tirexTwo{} (ZS) & 0.6973 & 0.4781 & -- & -- & -- & -- \\
Toto-1.0 & 0.7501 & 0.5173 & 0.6980 & 0.5839 & 28.23 & 40.76 \\
\totoTwoSize{2.5B} & 0.6956 & 0.4759 & 0.6419 & 0.5394 & 32.54 & 44.42 \\
\totoTwoSize{1B} & 0.6992 & 0.4784 & 0.6448 & 0.5446 & 32.39 & 44.43 \\
\totoTwoSize{313M} & 0.7028 & 0.4814 & 0.6444 & 0.5423 & 32.10 & 44.23 \\
\totoTwoSize{22M} & 0.7188 & 0.4963 & 0.6703 & 0.5616 & 30.61 & 42.91 \\
\totoTwoSize{4M} & 0.7565 & 0.5242 & 0.6910 & 0.5825 & 28.04 & 40.71 \\
\bottomrule
\end{tabular}
\end{table*}

\FloatBarrier

\subsection{IND-Eval}
\label{app:ind-eval}

\paragraph{Benchmark.}
We additionally evaluate \timestral on IND-Eval~\citep{haugland2026indevalaligningevaluationsimulation}, a public benchmark for forecasting in complex physical systems.
It comprises 187 tasks covering applications in energy, manufacturing, and transport.
Compared with general-purpose forecasting benchmarks, it emphasises finer sampling resolutions and multivariate targets.

\paragraph{Evaluation.}
All IND-Eval results reported here come from our own evaluation runs, using publicly released checkpoints and inference code from the model authors' public repositories under the official benchmark protocol.
We re-evaluate \timesfmThree, \chronosTwo, and \tirexTwo, the three models studied by \citet{haugland2026indevalaligningevaluationsimulation}, and extend the comparison to \tzeroGamma, t0-beta, all five released \totoTwo sizes (4M, 22M, 313M, 1B, and 2.5B), and Granite-PatchTST-FM-r2, alongside all four \timestral sizes.
All models are evaluated zero-shot, without fine-tuning, using the official IND-Eval context--target windows and a context capped at 4,096 time steps.
We report \ac{nmase} and \ac{ncrps} relative to Seasonal Naive. Each task score divides the model's error by the corresponding error of Seasonal Naive; values below one indicate lower error.
We first take the geometric mean of task scores within each dataset, then the geometric mean of the resulting dataset scores.
This gives each dataset equal weight in the aggregate.

\begin{figure}[t]
    \centering
    \includegraphics[width=1.0\linewidth,page=4]{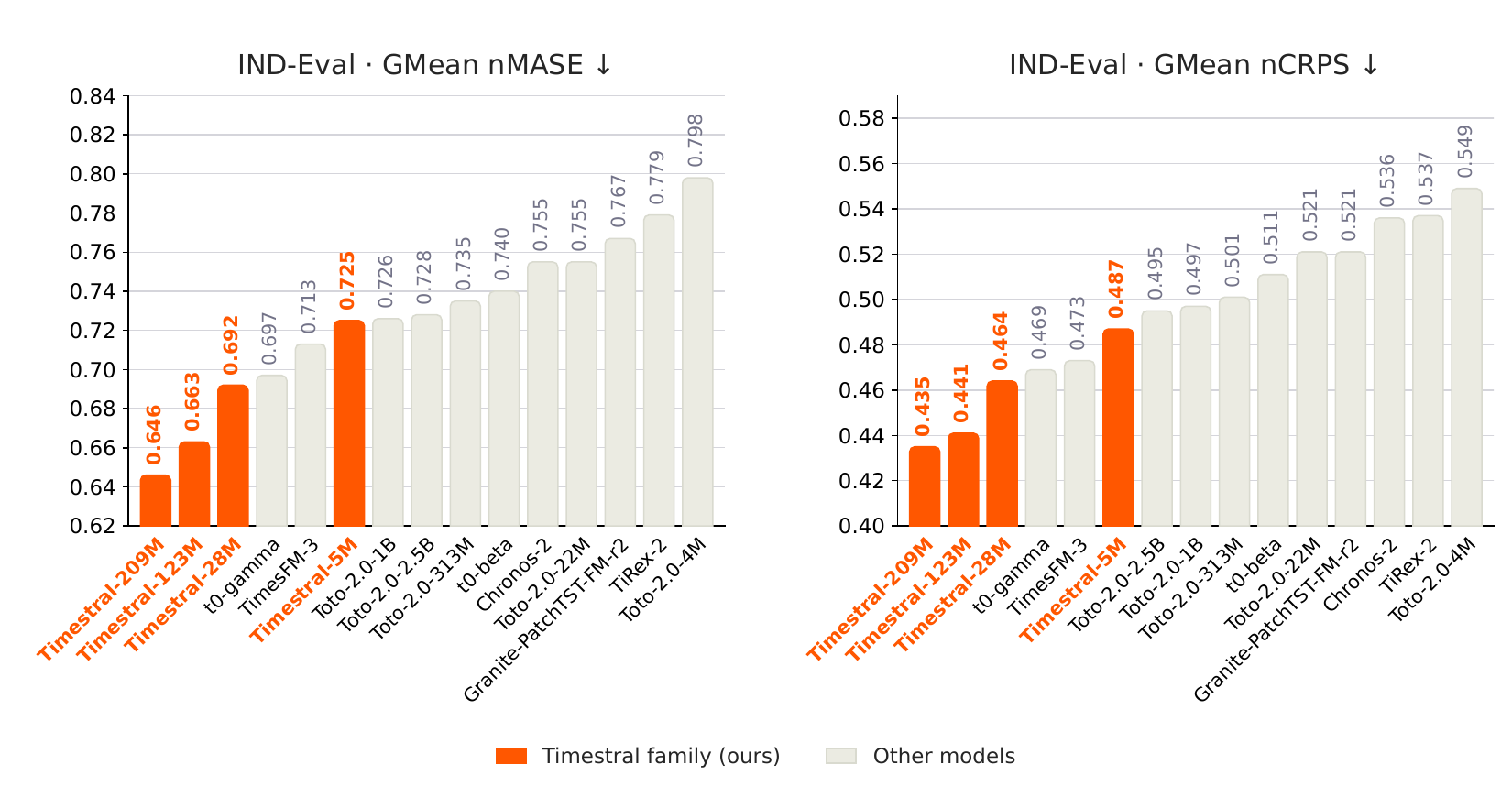}
    \caption{IND-Eval aggregate \ac{nmase} (left) and \ac{ncrps} (right) for all four \timestral sizes and eleven competitors. 
    All results are from our evaluations using public checkpoints and inference implementations under the official benchmark protocol. 
    Both panels show the same models, ordered by their respective dataset-balanced geometric-mean error. Lower is better.}
    \label{fig:results:ind-eval}
\end{figure}

\paragraph{Aggregate results.}
\Cref{fig:results:ind-eval} compares aggregate errors for all four \timestral sizes and the eleven competitors evaluated here.
\timestralSize{209M} ranks first among the models compared, with aggregate \ac{nmase} of 0.646 and \ac{ncrps} of 0.435. Relative to \tzeroGamma, the strongest model outside the \timestral family on both metrics (0.697 and 0.469), this is about 7\% lower error. 
Accuracy improves at every increase in \timestral model size. 
\timestralSize{123M} obtains 0.663 \ac{nmase} and 0.441 \ac{ncrps}, and \timestralSize{28M} obtains 0.692 and 0.464, respectively; both remain ahead of every model outside the family on the two aggregates.

\clearpage
\subsection{Pairwise Win Rates}
\label{app:win-rates}

\begin{figure}[htbp]
    \centering
    \includegraphics[page=1,width=0.7\linewidth]{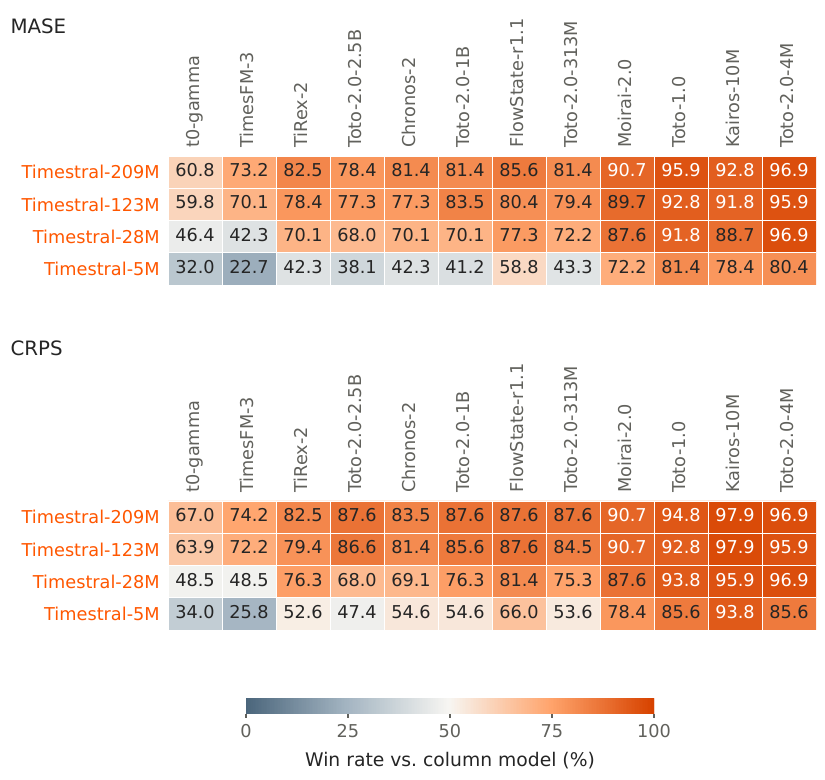}
    \caption{GIFT-Eval pairwise win rates for \acs{mase} (top) and \acs{crps} (bottom) across 97 dataset--frequency--horizon configurations. Rates above 50\% favour the row model. \acs{crps} uses the official nine-quantile approximation; TiRex-2 is the pretrained submission. Both panels order comparators by increasing aggregate \acs{mase}.}
    \label{fig:gift-win-rates}
\end{figure}

\begin{figure}[htbp]
    \centering
    \includegraphics[page=2,width=0.7\linewidth]{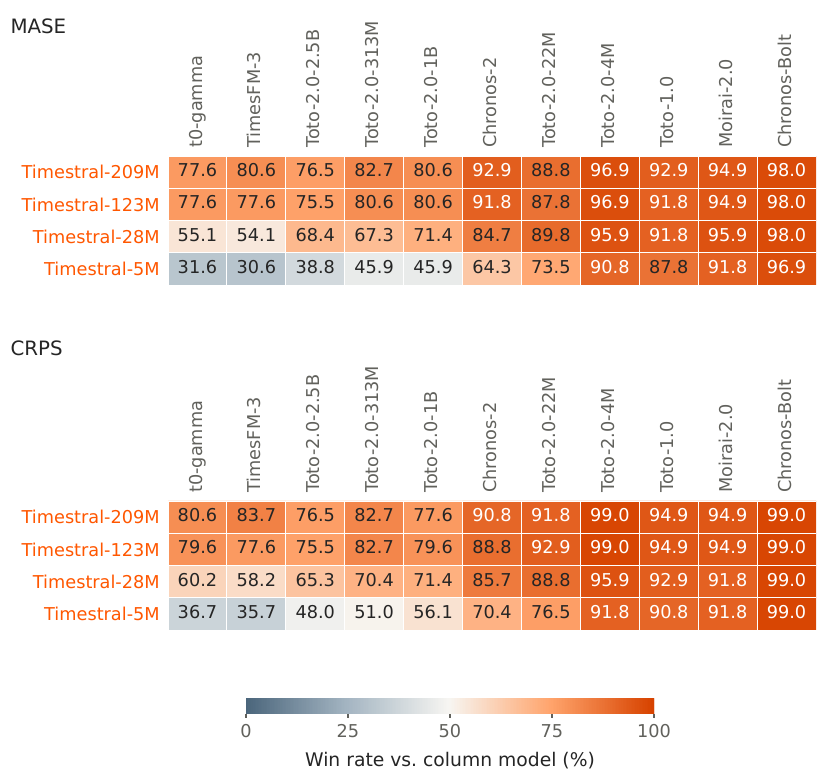}
    \caption{TIME pairwise win rates for \acs{mase} (top) and \acs{crps} (bottom) across 98 dataset--frequency--horizon configurations. Task errors follow the official leaderboard, averaging over series, forecast windows, and variates. Configurations are equally weighted, with ties counted as half a win; rates above 50\% favour the row model. Both panels order comparators by increasing aggregate \acs{mase}.}
    \label{fig:time-win-rates}
\end{figure}

\begin{figure}[htbp]
    \centering
    \includegraphics[page=3,width=0.7\linewidth]{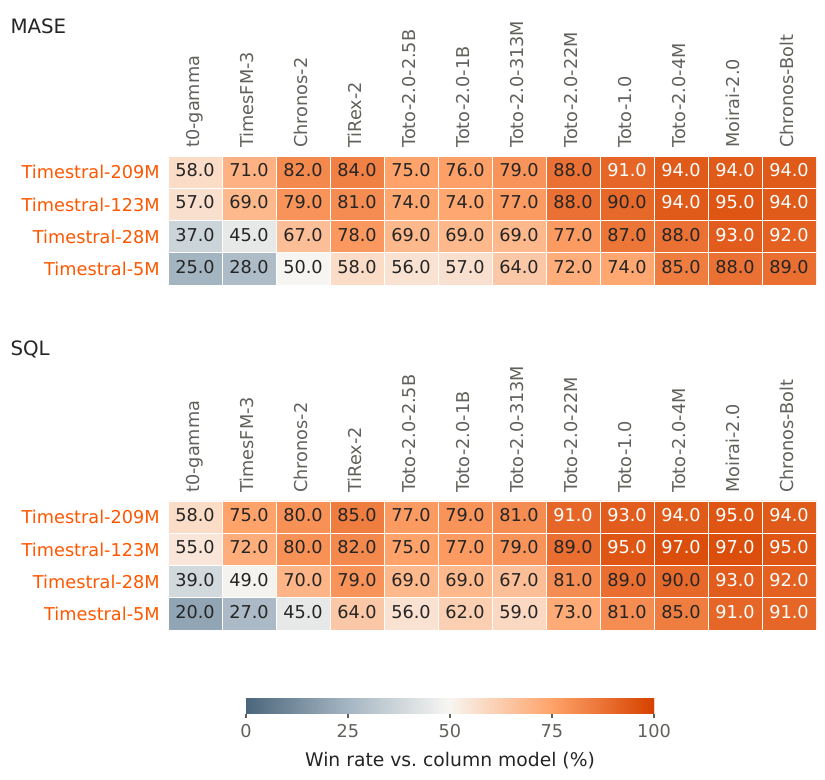}
    \caption{fev-bench pairwise win rates for \acs{mase} (top) and \acs{sql} (bottom) across 100 official tasks. \acs{sql} scales quantile loss by the context's mean absolute seasonal error. Tasks are equally weighted, with ties counted as half a win; rates above 50\% favour the row model. Both panels order comparators by decreasing MASE skill score.}
    \label{fig:fev-win-rates}
\end{figure}

\begin{figure}[htbp]
    \centering
    \includegraphics[page=4,width=0.7\linewidth]{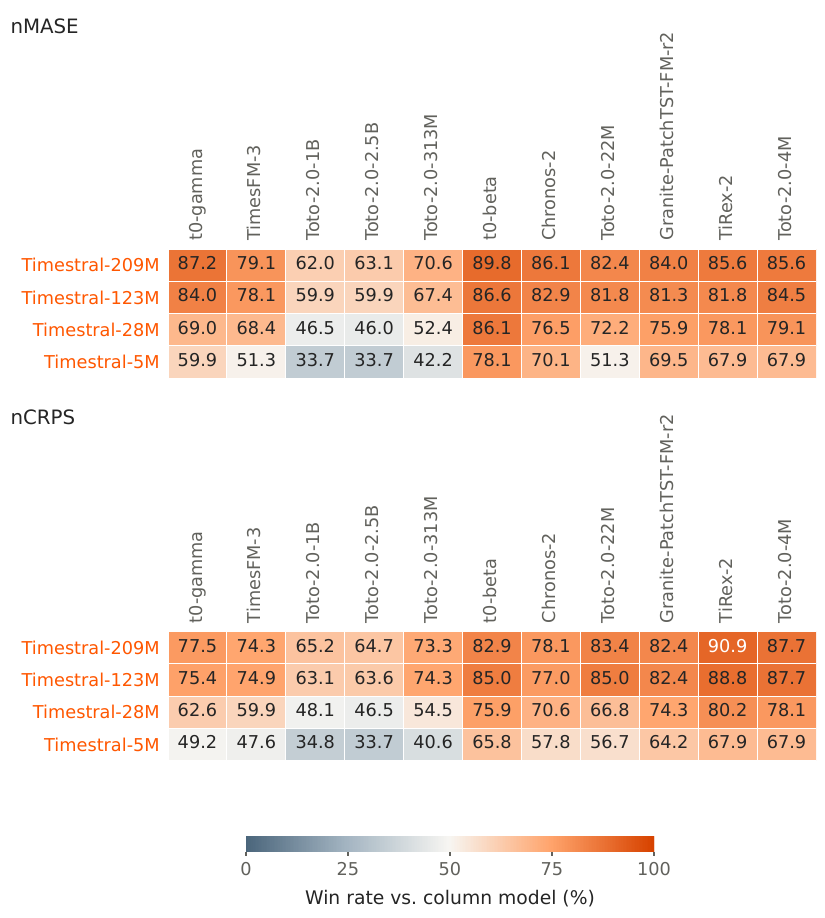}
    \caption{IND-Eval pairwise win rates for \acs{nmase} (top) and \acs{ncrps} (bottom) across 187 forecasting tasks: all four \timestral sizes versus the eleven competitors in \cref{fig:results:ind-eval}. Errors come from our evaluations under the official protocol. Tasks are equally weighted, with ties counted as half a win; aggregate errors instead weight datasets equally. Rates above 50\% favour the row model. Both panels order comparators by increasing aggregate \acs{nmase}.}
    \label{fig:ind-win-rates}
\end{figure}

Aggregate errors summarise overall forecasting accuracy, while pairwise win rates measure how often one model outperforms another across tasks.
\Cref{fig:gift-win-rates,fig:time-win-rates,fig:fev-win-rates,fig:ind-win-rates} compare all four released \timestral sizes with the selected competitors on each benchmark.
Each cell reports the row model's win rate against the column model across the benchmark's tasks, with lower error counting as a win and exact ties as half a win, following the fev definition~\citep{shchur2025fevbench}.
Errors are compared before rounding, and all tasks receive equal weight.

We follow the official evaluation protocols for each benchmark; IND-Eval task errors come from our own evaluation runs described in \cref{app:ind-eval}.
For fev-bench, win rates use task errors rather than aggregate skill scores.
\timestralSize{209M} wins on a majority of tasks against every selected competitor on both metrics of each benchmark.
\timestralSize{123M} shows the same pattern on all four benchmarks.

\end{document}